\documentclass[twocolumn,english,final]{IEEEtran}
\usepackage[T1]{fontenc}
\usepackage[latin9]{inputenc}
\usepackage{textcomp}
\usepackage{url}
\usepackage{amsmath}
\usepackage{amssymb}
\usepackage{graphicx}

\makeatletter

\providecommand{\tabularnewline}{\\}
\newcommand{\lyxdot}{.}

\usepackage{flushend}

\makeatother

\usepackage{babel}
\begin{document}

\title{Stochastic Texture Difference for Scale-Dependent Data Analysis}

\author{Nicolas Brodu, Hussein Yahia}
\maketitle
\begin{abstract}
This article introduces the Stochastic Texture Difference method for
analyzing data at prescribed spatial and value scales. This method
relies on constrained random walks around each pixel, describing how
nearby image values typically evolve on each side of this pixel. Textures
are represented as probability distributions of such random walks,
so a texture difference operator is statistically defined as a distance
between these distributions in a suitable reproducing kernel Hilbert
space. The method is thus not limited to scalar pixel values: any
data type for which a kernel is available may be considered, from
color triplets and multispectral vector data to strings, graphs, and
more. By adjusting the size of the neighborhoods that are compared,
the method is implicitly scale-dependent. It is also able to focus
on either small changes or large gradients. We demonstrate how it
can be used to infer spatial and data value characteristic scales
in measured signals and natural images.
\end{abstract}

\section{Introduction}

\subsection{What this work is about}

In this paper, we present a local algorithm, a low-level and scale-dependent
data analysis that is able to identify characteristic scales in data
sets, for example (but not limited to) images. The main idea is to
statistically characterize how data values evolve when exploring each
side of a given pixel, then compare the statistical properties of
each side. We exploit random walks on images but in a different setting
than \cite{random_walkers}. Our method resembles the graph walks
in \cite{graph_walks}, except that we additionally introduce a statistical
comparison operator, that our method is adapted to images, directional,
accounts for neighborhood information and is intrinsically scale-dependent.
Indeed, exploring the neighborhood of each pixel implies setting two
scales: the spatial extent of that exploration, and the sensitivity
to pixel value differences along each explored path. Once these are
set, we quantify the difference between each side, and repeat the
operation in other orientations around a pixel. This gives, for each
pixel and each prescribed set of (spatial, data) scales, a measure
of how that pixel is a transition between distinct statistical properties
on each of its sides. Depending on how these scales are set, we may
for example get a new kind of low-level boundary detector, sensitive
to either small data changes or large gradients, and at different
resolutions. The method can be used to infer characteristic scales
in natural data, these at which the boundaries capture most of the
information. The method is not restricted to images, it is readily
extendable to voxels or other spatially extended data (e.g. irregular
meshes). It is not restricted either to scalar (e.g. gray) valued
pixels: any data type for which a reproducing kernel is available
can be used (e.g. vector, strings, graphs). We present in particular
the case for color triplets in Section \ref{sec:Color-texture-differences}.

\subsection{What this work is not about}

Our goal is to provide a local algorithm, that works at pixel level,
and not to perform global image segmentation. The best algorithms
for segmentation account for texture information \cite{Ren12}, exploit
eigen-decompositions for identifying similar regions \cite{Shi00},
or improve their results with edge completion \cite{Maire08}. These
high-level operations are very efficient at finding transitions between
general zones in the image \cite{berkeley_segmentation}, but we want
a local quantifier that can be used to infer characteristic scales
within acquired data, which is a different problem from image segmentation.
It may be the case that our methodology becomes useful in a segmentation
context, but this topic is out of the scope of the present paper.

Similarly, our goal is not to perform high-level texture classification
\cite{zhang07}, but to formulate a new way to represent textures
and their differences (Section \ref{sec:Comparing-textures}). As
a quantifier for texture differences, our method might provide future
works with a new way to compare an unknown sample with a reference
database. However, it remains to be seen how efficient that would
be compared to well-established alternatives. In particular, we do
not decompose data on a basis of functions, such as wavelets \cite{Khouzani05},
curvelets \cite{Starck02}, bandelets \cite{LePennec2005} and other
variants \cite{Do05}. Neither do we adapt the decomposition basis
using a dictionary \cite{Mairal08} or define elementary textons \cite{Malik01}
for description and classification purposes. We compare distribution
of probabilities, which we identify with the textures. It may be that
our method ultimately helps in defining new optimal decomposition
of images on some basis of elementary textures, but that is another
issue, which falls out of the scope of this article.

\subsection{Comparison with related approaches}

Multi-scale analysis can either be performed by exploiting relations
and patterns across scales, or by defining features per scale, and
then investigating how these evolve as the scale varies. The first
category includes all wavelet-based methods mentioned above, as well
as more advanced techniques like multifractal analysis \cite{Badri14,Xia06}.
By definition, methods in this first family do not allow finding characteristic
scales, they exploit relations between scales, for example to build
texture descriptors. The second category, a scale-dependent analysis
which is then repeated at multiple scales, is prominently represented
in the field of image processing by space-scale decompositions \cite{Lindeberg94}.
This way, edges are found at different levels of details \cite{Lindeberg98},
hopefully allowing the user to remove all extraneous changes and focus
on important object boundaries \cite{Stella2005}. Any scale dependent
analysis might be used in this second family, and in particular pixel
connectivity, which is sometimes used to obtain multiscale descriptions,
see \cite{BragaNeto04} for a review. Our method falls in the second
category, but with a completely new feature involving sets of random
paths. It allows to specify both the spatial scale and the data scale
for the analysis, and to look for optima in this two-dimentional parameter
space which are presumably characteristic of the data, see Section
\ref{sec:Multi-scale-analysis}.

We then use statistics to quantify how data values evolve in a pixel
neighborhood at these prescribed spatial and data scales. \cite{Konishi03}
also uses statistics to compare data distributions at multiple scales,
but with a different framework and for classification purposes. We
use random walks in order to explore the neighborhoods. Although we
use a Markov Random Field (MRF) approach in order to fix the random
walk transition probabilities, our approach is fundamentally different
from the typical use of MRF in image processing (see \cite{Schmidt10}
for a review). We use the MRF solely as a intermediate step to calibrate
the random walks with a spatial scale, and not as a model for pixel
values, or for segmentation \cite{random_walkers,Liang06}. Anisotropic
diffusion methods, from the Perona-Malik model \cite{Perona90} to
space-scale extensions with partial differential equations \cite{Petrovic04},
might seem related to this random walk approach. Indeed, gaussian
filtering could be seen as integrating pixel values along random paths
in the neighborhoods we use. However, we do not integrate along random
paths, but we rather exploit how pixel values evolve along these paths
in order to define a statistical distribution of these variations
in the neighborhood, which we then identify with texture information
(Section \ref{sub:Quantifying-texture-differences}). Our method can
thus be seen as complementary to diffusion. A notable multiscale extension
of anisotropic diffusion to vector data can be found in \cite{Dong06},
and it would be interesting to compare how introducing a reproducing
kernel in their setup complements our own technique.

Our method is applicable to non-scalar data, including non-vector
data, by means of reproducing kernels. Yet, we do not use kernel methods
as in machine learning of texture features \cite{Fernandez13,zhang07},
but as an indirect way to quantify discrepancies between probability
distributions, building on recent findings in statistics \cite{RKHS_proba_measures,mmd}.
Consequently, for the particular case of kernels on color spaces which
we demonstrate in Section \ref{sec:Color-texture-differences}, our
method is fundamentally different from current approaches in the domain
\cite{Fowlkes04}. Our hope is that this ability to work with non-scalar
data will make our method a prominent tool for the analysis of multispectral
and other kind of images. 

The stochastic texture discrepancy (STD) which we present can also
be seen as a kind of ``texture gradient'' (see Section \ref{sub:Selective-texture-erasing}).
Unlike related work, which use wavelet transforms \cite{Clerc02},
morphological operations \cite{Angulo07}, or combinations of both
\cite{Callaghan05} to define texture gradients, the STD is a norm
in a suitable reproducing kernel Hilbert space. Consequently, we expect
that it could be also incorporated in classic gradient-based edge-detection
algorithms \cite{Canny86}, or any other setup in which a proper norm
is required.

Section II presents the theory, starting with how texture is encoded
and Section III how texture information is compared on each side of
a pixel. Section IV gives a first application of the method, how it
can detect texture boundaries accouting for scale information. Section
V builds on this and presents a fully automated method for detecting
characteristic scales in acquired data (including, as in this paper,
images). Section VI demonstrates that our method deals equally well
with non-scalar data, including, in that application case, color triplets.
Color texture difference is shown to better detect boundaries and
highlight small details than alternatives.

\section{\label{sec:texinfo}Encoding texture information}

\begin{figure*}
\includegraphics[width=1\textwidth]{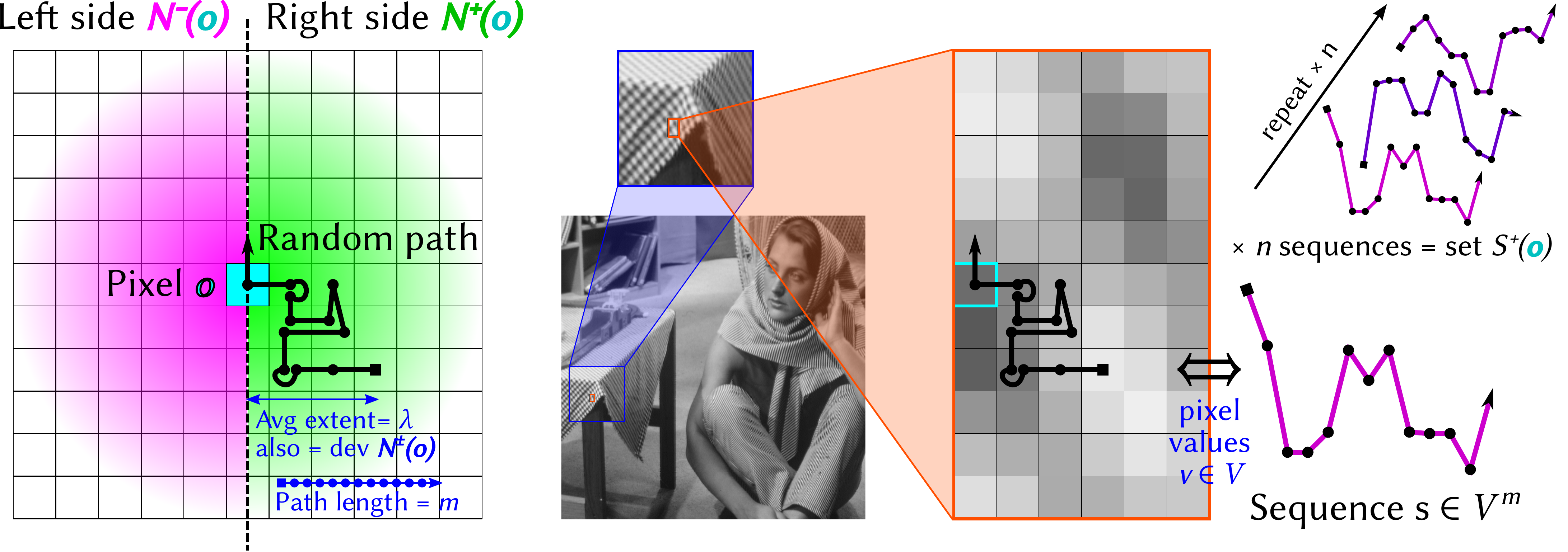}

\caption{\label{fig:Method}General method for generating stochastic texture
descriptions. The case for a left/right split is presented, but other
orientations with any angle are processed in exactly the same way.
Notations and a step by step description are given in the main text.}
\end{figure*}

In the following, bold mathematical symbols denote pixels, plain symbols
are values associated to these pixels. Capital mathematical symbols
(resp. bold, plain) represent sets (of resp. pixels, values). Mathematical
spaces are calligraphied.

\subsection{Overview}

Consider a pixel $\boldsymbol{o}$. An orthonormal basis $\left(\vec{x},\vec{y}\right)$
in the image plane being fixed (for instance the one aligned with
the boundaries of the image), we consider the affine basis $\left(\boldsymbol{o},\vec{x},\vec{y}\right)$
with the origin at that pixel center. We want to compare how the texture
on the half-plane $\{x>0\}$ is statistically different from the texture
on the other side $\{x<0\}$ (Fig.~\ref{fig:Method}). A length scale
is needed to characterize the extent of the texture comparison within
each half-plane. Let $\lambda$ be this spatial length scale. Noting
$\boldsymbol{N}^{\pm}(\boldsymbol{o})$ neighborhoods of pixels within
the two half-planes, consider the probability densities defined on
$\boldsymbol{N}^{\pm}(\boldsymbol{o})$ by $p^{\pm}(x,y)=C\cdot\exp\left(-\frac{x^{2}+y^{2}}{2\lambda^{2}}\right)$
for $\pm x\ge0$ and $0$ otherwise, where $C$ is a normalizing constant.
The texture characterization we propose amounts to:

A: monitoring pixel values taken along $n$ independent random paths
in each of these neighborhoods $\boldsymbol{N}^{\pm}(\boldsymbol{o})$;

B: in a way that the probability to visit a pixel $\boldsymbol{a}\in\boldsymbol{N}^{\pm}(\boldsymbol{o})$
is given by $p^{\pm}(\boldsymbol{a})=\iint_{x,y\in\boldsymbol{a}}p^{\pm}(x,y)\mbox{d}x\mbox{d}y$;

C: and that the average spatial extent of a random path is $\lambda$.

As a result of this process, two sets $S^{\pm}(\boldsymbol{o})$ are
obtained for each pixel $\boldsymbol{o}$ along each pair of opposite
directions%
\footnote{Pixels along the boundary line may appear in each set, as shown in
Fig.~\ref{fig:Method}. In this vertical line example pixels fall
on either side with equal probability, but the method is applicable
for lines at any angle, in which case $p^{-}(\boldsymbol{a})\neq p^{+}(\boldsymbol{a})$,
both being determined by the area of the pixel falling on either side
of the line as specified by $p^{\pm}(\boldsymbol{a})=\iint_{x,y\in\boldsymbol{a}}p^{\pm}(x,y)\mbox{d}x\mbox{d}y$.%
}. These sets each contain $n$ random sequences of pixel values $\left(v\left(\boldsymbol{a}_{i}\right)\right)_{i=1\ldots m}$,
with $v(\boldsymbol{a})$ the image information (gray scale, RGB,
etc.) at pixel $\boldsymbol{a}$. A set of paths thus contains the
statistical description of a texture on the corresponding side of
a pixel $\boldsymbol{o}$, at a prescribed characteristic spatial
scale of $\lambda$.

Each of the above points is detailed separately in the following subsections.

\subsection{Generating random paths}

Each pixel location around $\boldsymbol{o}$ is considered to be a
distinct state of a Markov chain. Transitions are allowed to the 4
neighborhood pixels, together with self-transitions. A random pixel
in $\boldsymbol{a}_{1}\in\boldsymbol{N}^{\pm}(\boldsymbol{o})$ is
chosen as the starting point, with a probability given by the limit
distribution $p^{\pm}\left(\boldsymbol{a}_{1}\right)$.

The Markov chain is then run for $m$ steps for generating the pixel
sequence $\left(\boldsymbol{a}_{i}\right)_{i=1\ldots m}$. The values
of these pixels are then taken along this path, so that $s=\left(s_{i}=v\left(\boldsymbol{a}_{i}\right)\right)_{i=1\ldots m}\in S^{\pm}(\boldsymbol{o})$
is a length $m$ excerpt of the texture in $\boldsymbol{N}^{\pm}(\boldsymbol{o})$.
The process is repeated $n$ times to get the full set $S^{\pm}(\boldsymbol{o})$,
in each direction, which is assumed to encode the full texture information
provided $n$ is large enough (Fig.~\ref{fig:Method}).

\subsection{Building the Markov chain}

For each pixel $\boldsymbol{a}$, hence each state of the Markov chain,
we first compute $p^{\pm}(\boldsymbol{a})$. This is the probability
of reaching this state in the limit distribution of the Markov chain.
We now need to define a set of transitions that leads to this limit
distribution.

Let $p^{\pm}(\boldsymbol{a}\rightarrow\boldsymbol{b})$ be the transition
probability from state/pixel $\boldsymbol{a}$ to state/pixel $\boldsymbol{b}$.
There are at most 5 non-null such probability transitions, from $\boldsymbol{a}$
to its neighborhood pixels $\boldsymbol{b}_{1\ldots4}$ and to itself
$\boldsymbol{b}_{5}=\boldsymbol{a}$. The Markov chain is consistent
with the limit distribution when:

\begin{equation}
p^{\pm}(\boldsymbol{a})=\sum_{i=1}^{5}p^{\pm}(\boldsymbol{b}_{i})p^{\pm}(\boldsymbol{b}_{i}\rightarrow\boldsymbol{a})\label{eq:Markov_condition_1}
\end{equation}

while respecting the constraint

\begin{equation}
\sum_{i=1}^{5}p^{\pm}(\boldsymbol{a}\rightarrow\boldsymbol{b}_{i})=1\label{eq:Markov_condition_2}
\end{equation}

The unknowns in this problem are the transition probabilities for
every pair $\boldsymbol{a},\boldsymbol{b}$. We start by assigning
initial transition probabilities $p_{ini}^{\pm}\left(\boldsymbol{a}\rightarrow\boldsymbol{b}_{i}\right)=p^{\pm}(\boldsymbol{b}_{i})/\sum_{i=1}^{5}p^{\pm}(\boldsymbol{b}_{i})$.
This, of course, does not provide the correct transition probabilities,
just a starting point. We then run an iterative Levenberg-Marquardt
optimisation scheme in order to solve%
\footnote{We use the solver \cite{Ceres} and reach a typical sum of squared
error the order of $10^{-30}$ for these conditions for both equations,
with a maximum of $10^{-14}$ for the smallest neighborhoods. We nevertheless
renormalize condition Eq.\ref{eq:Markov_condition_2} to ensure a
strict equality.%
} Eq.\ref{eq:Markov_condition_1} and Eq.\ref{eq:Markov_condition_2}.

\subsection{Ensuring sequences are consistent with the spatial scale}

For each sequence length $m$, we note the average extent $e(x)$
in $x$ for generated sequences of this size: $e(x)=\left\langle \max_{i\leq m}\left(\bar{x}\left(\boldsymbol{a}_{i}\right)\right)-\min_{i\leq m}\left(\bar{x}\left(\boldsymbol{a}_{i}\right)\right)\right\rangle $,
with $\bar{x}\left(\boldsymbol{a}\right)$ denoting the $x$ coordinate
of the center%
\footnote{Or, equivalently, the mean of $x$ on the surface of pixel $\boldsymbol{a}$
in this case with centered pixels%
} of pixel $\boldsymbol{a}$. Averaging $e(x)$ is done numerically
over a large number of samples. We then select the value of $m$ that
gives a sequence of spatial extent $e(x)=\lambda$ on average. The
typical error on $\lambda$ is less than $0.01$ pixel for the neighborhoods
we use.

Building the Markov chain and defining $m$ needs to be done only
once for each neighborhood size $\lambda$ and for each considered
orientation of the neighborhoods $\boldsymbol{N}^{\pm}(\boldsymbol{o})$.
Distinct orientations lead to distinct $p^{\pm}(\boldsymbol{a})=\iint_{x,y\in\boldsymbol{a}}p^{\pm}(x,y)\mbox{d}x\mbox{d}y$
for every pixel $\boldsymbol{a}$ around $\boldsymbol{o}$. We limit
ourselves in this paper to straight and diagonal orientations, so
the densities $p^{\pm}(\boldsymbol{a})$ present a convenient symmetry
in each of these two cases, but any angle is possible. Once these
are precomputed, the same (straight, diagonal) Markov chains can be
applied to each pixel $\boldsymbol{o}$ in the image (boundary conditions
are dealt with in Section \ref{sub:Handling-missing-data}).

\section{\label{sec:Comparing-textures}Comparing textures}

The procedure described in the previous section is applied to each
pixel on an image, leading to 4 pairs of sets $S^{\pm}(\boldsymbol{o})$
for each pixel $\boldsymbol{o}$, corresponding to both straight and
diagonal directions. Each set comprises $n$ sequences $s=\left(s_{i}=v\left(\boldsymbol{a}_{i}\right)\right)_{i=1\ldots m}$
of length $m$, with $m$ chosen so that sequences extend spatially
to a characteristic scale $\lambda$ in the corresponding direction
on average. This section details how to quantify the difference $d\left(S^{-},S^{+}\right)$
between two sets $S^{\pm}(\boldsymbol{o})$, hence between textures
on opposite sides of a pixel.

\subsection{\label{sub:Comparing-two-sequences}Comparing two sequences of pixel
values}

Let $s=\left(s_{i}=v\left(\boldsymbol{a}_{i}\right)\right)_{i=1\ldots m}$
be a sequence as defined above. Let $\mathcal{V}$ be the space taken
by pixel values $v(\boldsymbol{a})\in\mathcal{V}$, so that $s\in\mathcal{V}^{m}$.
Consider a kernel $k:\,\mathcal{V}^{m}\times\mathcal{V}^{m}\rightarrow\mathbb{R}$,
such that:

\textendash{} $k\left(s,\cdot\right)$ is a function, in the Hilbert
space $\mathcal{H}$ of functions from $\mathcal{V}^{m}$ to $\mathbb{R}$.

\textendash{} For any function $f\in\mathcal{H}$ and for all $s\in\mathcal{V}^{m}$,
the reproducing property holds: $\left\langle f,k(s,\cdot)\right\rangle _{\mathcal{H}}=f(s)$
with $\left\langle \cdot,\cdot\right\rangle _{\mathcal{H}}$ the inner
product in $\mathcal{H}$.

\textendash{} $\mathrm{Span}\left\{ k(s,\cdot)\right\} _{s\in\mathcal{V}^{m}}$
is dense in $\mathcal{H}$

As a special case, the kernel value $k\left(s^{-},s^{+}\right)=\left\langle k\left(s^{-},\cdot\right),k\left(s^{+},\cdot\right)\right\rangle _{\mathcal{H}}$
gives the similarity between $s^{-}$ and $s^{+}$. As is customary,
we normalize $k$ such that $k\left(s,s\right)=1$ and $k\left(s,t\neq s\right)<1$
for all $s,t\in\mathcal{V}^{m}$.

This classic reproducing kernel Hilbert space (RKHS) representation
lets us consider any pixel value: scalar (e.g. gray scale, reflectance),
vector (e.g. RGB, hyperspectral components), or in fact any kind of
data for which a kernel is available (e.g. graph, strings). Indeed,
once a kernel is available for an element $s_{i}\in\mathcal{V}$,
it is always possible to use the product kernel for the sequence $s\in\mathcal{V}^{m}$,
hence to compute the similarity between two sequences. The method
we present is thus generic, and applicable to wide range of ``images'',
broadly defined as spatially extended data. In particular, one way
to handle color textures is presented in Section \ref{sec:Color-texture-differences}.
Similarly, definitions in Section \ref{sec:texinfo} are easily extended
to more than 2 spatial dimensions if needed (e.g. Markov chains transitions
to the 6 voxel neighbors in 3 dimensions), or even to non-regular
meshes (by computing $p^{\pm}(\boldsymbol{a})$ for each mesh cell
and transitions as above), or graphs (provided a meaninful $p^{\pm}(\boldsymbol{a})$
can be provided for each node $\boldsymbol{a}$).

\subsection{\label{sub:Quantifying-texture-differences}Quantifying texture differences}

The random sequences defined above can be seen as samples of an underlying
probability distribution over $\mathcal{V}^{m}$, which we identify
with the texture information. It is possible that visually distinct
patterns lead to the same distribution over $\mathcal{V}^{m}$, as
we offer no uniqueness proof, and especially with $n<\infty$. However,
the same visual ambiguity exists for other and widely used texture
representation methods, for example when retaining only a small number
of components in a suitable decomposition basis (\cite{Starck02,LePennec2005,Shi00}).
We make no claim on the superiority of our method in this respect,
but it certainly offers a very different characterization of textures,
complementary to other approaches.

Let us thus identify a texture as a probability distribution $P(s)$
over $\mathcal{V}^{m}$. In this view, the sets $S^{\pm}(\boldsymbol{o})$
are collections of observed samples for the textures in the neighborhoods
$\boldsymbol{N}^{\pm}(\boldsymbol{o})$. Quantifying the texture difference
$d\left(S^{-},S^{+}\right)$ amounts to performing a two-sample test
for similar distributions, with similarity defined by a metric on
probability distributions. Using the above RKHS representation, we
exploit recent litterature \cite{RKHS_proba_measures,mmd} to propose
a consistent estimator for $d\left(S^{-},S^{+}\right)$ that works
reliably even for small number of samples $n$, irrespectively of
the dimension of $\mathcal{V}$ and $\mathcal{V}^{m}$:

\textendash{} For a distribution $P$, the average map $\mu_{P}\in\mathcal{H}$
is given by $\mu_{P}=E_{P}\left[k(s,\cdot)\right]$.

\textendash{} An estimator $\hat{\mu}_{S}$ is easily obtained from
the sample set $S$, in the form $\hat{\mu}_{S}=\frac{1}{n}\sum_{j=1}^{n}k\left(s^{j},\cdot\right)$,
where $s^{j}$ denotes the $j$-th member of the set $S$.

\textendash{} Let $P$ and $Q$ be two distributions over $\mathcal{V}^{m}$.
If the kernel is characteristic, then $P=Q$ if and only if%
\footnote{Technically, this is only true up to an arbitrary set with null probability.
But such sets cannot be observed in practice in sampled data, hence
they can be safely ignored.%
} $\left\Vert \mu_{P}-\mu_{Q}\right\Vert _{\mathcal{H}}=0$. If $P\neq Q$,
the norm directly quantifies the discrepancy between these distributions.

Thus, the texture difference $d\left(S^{-},S^{+}\right)$ can be estimated
with:

\begin{gather}
d^{2}\left(S^{-},S^{+}\right)=\left\Vert \hat{\mu}_{S^{+}}-\hat{\mu}_{S^{-}}\right\Vert _{\mathcal{H}}^{2}\label{eq:MMD}\\
=\left\langle \hat{\mu}_{S^{+}}-\hat{\mu}_{S^{-}},\hat{\mu}_{S^{+}}-\hat{\mu}_{S^{-}}\right\rangle _{\mathcal{H}}\nonumber \\
=\frac{1}{n^{2}}\sum_{j=1}^{n}\sum_{k=1}^{n}\left(k\left(s_{+}^{j},s_{+}^{k}\right)+k\left(s_{-}^{j},s_{-}^{k}\right)-2k\left(s_{+}^{j},s_{-}^{k}\right)\right)\nonumber 
\end{gather}

Thanks to being a consistent estimator \cite{mmd} for a norm in a
suitable space, the operation we propose behaves very much like a
pixel-based gradient norm, but on textures differences instead of
pixel values, realizing of form of ``texture gradient'' (See Section
\ref{sub:Selective-texture-erasing}).

We define an overall Stochastic Texture Discrepancy $\textrm{STD}(\boldsymbol{o})$
for each pixel $\boldsymbol{o}$ by considering the norm in the product
space $\bigotimes_{\theta}\mathcal{H}_{\theta}$ over all directions
$\theta$, normalized by the number of directions, such that

\begin{equation}
\textrm{STD}(\boldsymbol{o})^{2}=\frac{1}{4}\sum_{\theta=1}^{4}d_{\theta}^{2}\left(S_{\theta}^{-},S_{\theta}^{+}\right)\label{eq:pixel_texture_discrepancy}
\end{equation}

with $\theta$ indexing in this paper the horizontal, vertical and
the two diagonal pairs of sets.

The whole procedure is easily generalizable to higher dimensions (e.g.
voxels), to arbitrary angles $\theta$ (by computing the Markov chain
for the limit distribution of the rotated neighborhood), to anisotropic
neighborhoods, etc.

\subsection{\label{sub:Choosing_kappa}Choosing an appropriate data scale}

In the results presented in the next sections, we use the inverse
quadratic kernel for scalar valued pixels:

\begin{equation}
k\left(s,t\right)=1/\left(1+\frac{1}{m}\sum_{i=1}^{m}\left(s_{i}/\kappa-t_{i}/\kappa\right)^{2}\right)\label{eq:inv_quad_kernel}
\end{equation}

This kernel is characteristic \cite{RKHS_proba_measures}, and we
found that it performs as well as the Gaussian kernel, but it is much
faster to compute. We have normalized the sum of squares by $m$,
so that straight and diagonal computations have comparable kernel
values. The parameter $\kappa$ sets the scale at which the interesting
dynamics occur in the observed signal values, and should be set a
priori by the user or a posteriori in order to optimize some objective
criterion. In Section \ref{sub:benchmark}, we use the reconstruction
Peak Signal to Noise Ratio (PSNR) as a measure of accuracy for each
value of $\lambda$ and $\kappa$. Other applications might have different
criteria, for example a classification rate in a machine learning
scenario. The important point is that the optimal spatial and data
scales should be set a posteriori with such criteria, as it is not
guaranteed that maximal PSNR (i.e. minimal squared reconstruction
error) matches the optimal criteria for every application (e.g. classification).
Moreover, in most images, some spatial variations for the optimal
$\kappa$ are expected, since not every part of the same image reflect
the same object. Hopefully, if so desired, our method can be applied
on arbitrarily shaped sub-parts of such images simply by masking out
other parts as missing data, as detailed in the next section.

\subsection{\label{sub:Handling-missing-data}Handling missing data and image
boundaries}

\begin{figure*}
\includegraphics[width=0.33\textwidth]{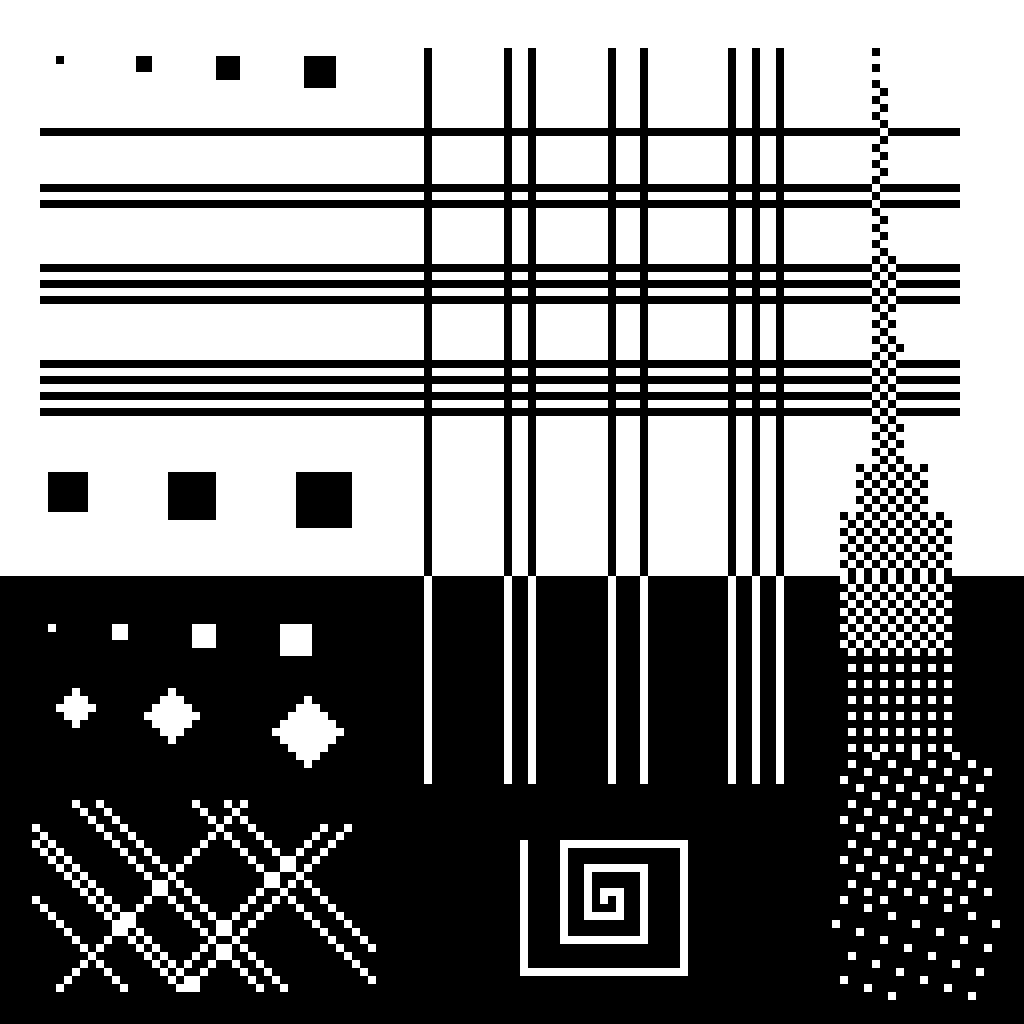}\hfill{}\includegraphics[width=0.33\textwidth]{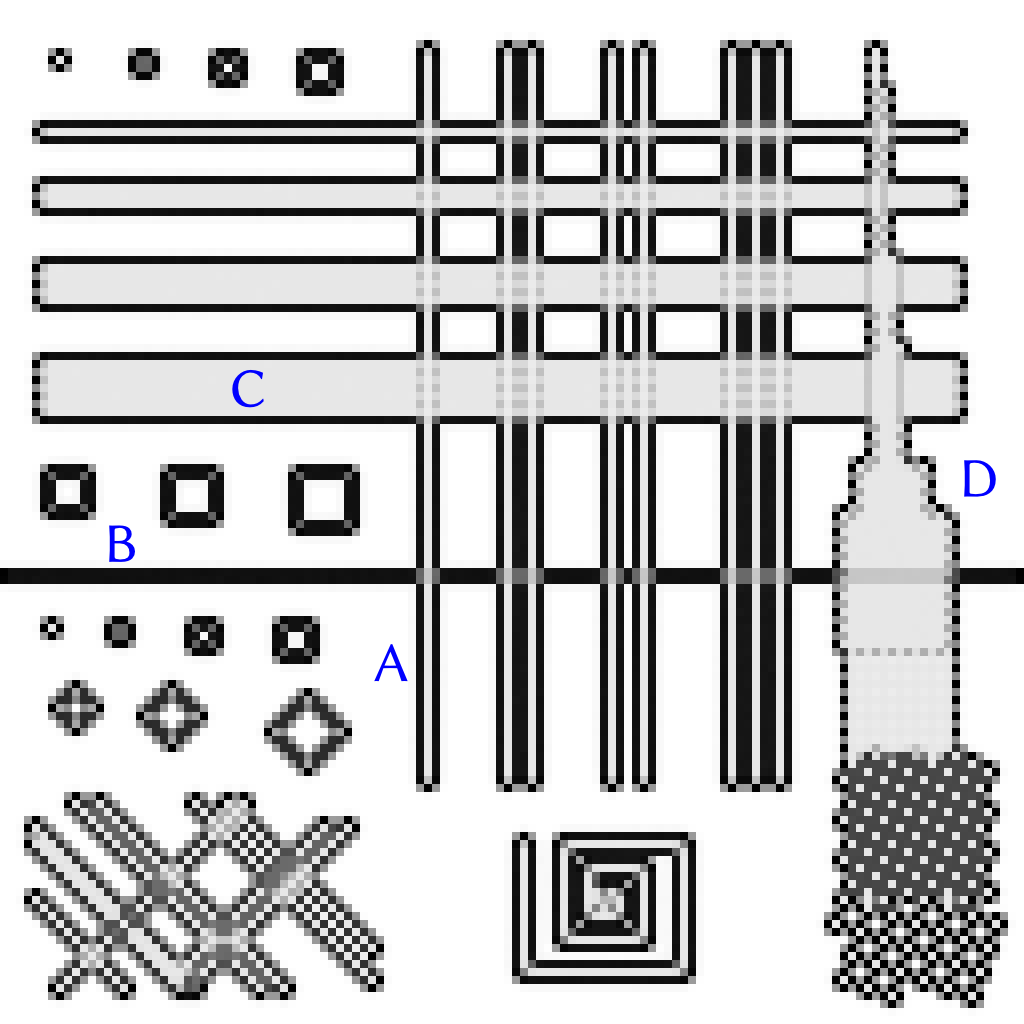}\hfill{}\includegraphics[width=0.33\textwidth]{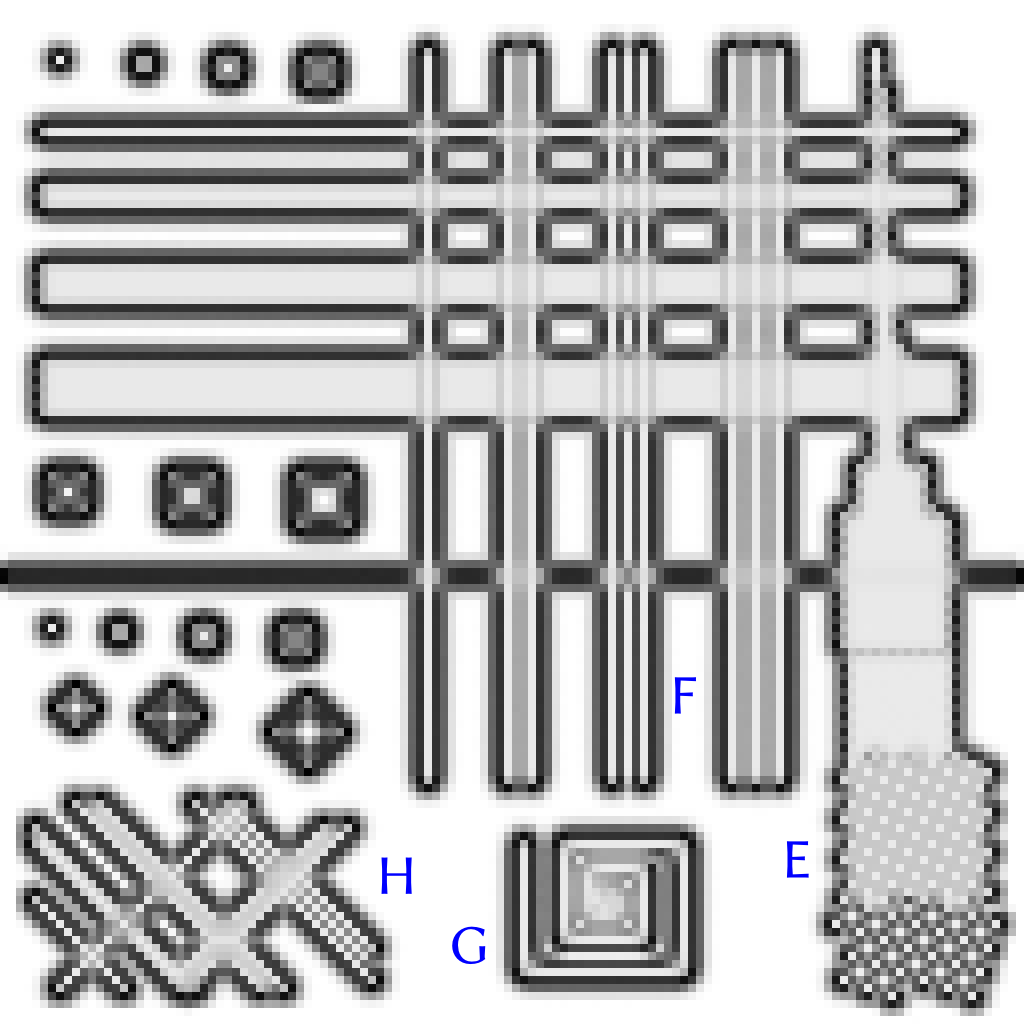}

\caption{\label{fig:refpics}Reference black and white picture (left), stochastic
texture differences at spatial scales of $\lambda=1$ (center) and
$\lambda=2$ pixels (right), both with value scale $\kappa=0.25$,
averaged over 10 independent computations with $n=500$. White stands
for a null difference, black for maximum discrepancy. Capital letters
markers are commented in the main text.}
\end{figure*}

Some images, like the satellite monitoring of sea surface temperature
in Fig.~\ref{fig:Sea-surface-temperature.}, present missing data
for some pixels, in this case corresponding to land masses. In that
case, entries in some sequences $\left\{ s_{i}\right\} _{i\leq m}$
may be missing. We deal with the issue by modifying the kernel to
maintain a normalized average over the remaining entries in these
sequences:

\begin{equation}
k\left(s,t\right)=1/\left(1+\frac{1}{\left|C\right|}\sum_{i\in C}\left(s_{i}/\kappa-t_{i}/\kappa\right)^{2}\right)\label{eq:kernel_with_NaN}
\end{equation}

where $C$ is the set of common indices where both sequences have
valid entries. When $C=\emptyset$, the kernel value cannot be computed.
In that case, we average over remaining kernel evaluations, separately
in each term of Eq. \ref{eq:MMD}. When $d\left(S^{-},S^{+}\right)$
cannot be evaluated in some directions $S^{\pm}$, then we average
over remaining directions in Eq. \ref{eq:pixel_texture_discrepancy}
to define the texture discrepancy at that pixel. When all directions
are invalid, for example in zones of missing data larger than the
neighborhood $\boldsymbol{N}^{\pm}(\boldsymbol{o})$, we set the texture
discrepancy to Not a Number, indicating missing information%
\footnote{STD values for missing data are also produced within the range of
the neighborhood of a valid data point, albeit with reduced precision
as the distance from that point increases. In practice, we ignore
these STD values and retain only points where the original data was
defined.%
}. When a large zone of missing data is present, then some directions
have to be ignored and textures are compared in the remaining directions.
This process adaptatively handles contours of missing data, discarding
automatically directions toward missing data and considering only
directions tangent to the large invalid zone. 

This effect is exploited to deal with pixels at the boundary of an
image. Standard techniques for handling such boundaries involve extending
border pixel values, mirroring, or setting outside pixel values to
zero. However, these methods are only needed to work around the inability
of an image processing algorithm to deal with missing data. Since
we can, the natural choice for pixels outside the image is to consider
them missing. Thanks to the above method, computations at an image
border are performed in exactly the same way as for any pixel within
the image. Textures to the very border of the image are taken into
account and, importantly, without introducing spurious discrepancies
(such as setting outside pixels to null would) or spurious similarities
(such as mirroring or extending border pixel values would).

\section{A scale-dependent texture discrepancy detector}

\subsection{Demonstration on a synthetic test picture}

The synthetic image of Fig.~\ref{fig:refpics} left was designed
to demonstrate the behavior of the algorithm on elementary patterns
such as one-pixel width lines and chess boards. Labels in Fig.~\ref{fig:refpics}
(middle, right) are added for clarity of the discussion. \textbf{A}:
points within thin (isolated) 1-pixel lines have the same neighborhood
on each side of the line. Points next to these lines have a maximal
difference between neighborhoods at 1-pixel scale, resulting in detected
edges on each side of the line. At scale 2, points two pixels away
also have asymmetric neighborhood, although less than the nearby pixels,
hence localization of the maxima is preserved between scales $\lambda=1$
and $\lambda=2$ (for small $\kappa$ values, see Section \ref{sub:Choosing_kappa}).
\textbf{B}: sharp black/white transitions result in two-pixel wide
boundaries, one pixel on each side of the transition, with weaker
contributions further away at scale $\lambda=2$. \textbf{C}: lines
separated by 1 pixel are fully recognized as being part of the same
texture at 1-pixel scale, hence disappear in the stochastic texture
difference (STD) maps at scales $\lambda=1$ and $\lambda=2$ . These
STD values are nevertheless not null, unlike that of the uniform background,
hence these patterns are grayed out instead of blanked out. Boundaries
around these texture regions are detected, as in the isolated line
case. \textbf{D}: chessboard patterns are also recognized at both
scales, together with surrounding edges. The difference between the
chessboard pattern and the spaced-out squares below is weakly detected
in both cases, while the black/white inversion within the checkboard
pattern is only visible at $\lambda=1$. \textbf{E}: this repetitive
pattern is better recognized at spatial scale $\lambda=2$ and pixel
value scale $\kappa=0.25$ (the average value over a repeating block),
but its elements are too isolated at scale $\lambda=1$. The boundaries
of that texture are also found at scale $\lambda=2$. Pixels in the
region below are too far apart for $\lambda\leq2$. \textbf{F}: thin
lines separated by 2 pixels also become part of the same texture with
a surrounding edge at $\lambda=2$, while lines separated by 3 pixels
are isolated in both cases. \textbf{G}: The spiral center is a non-repeating
pattern of lines 1 pixel apart, which can be recognized at scale $\lambda=2$
(see the black branches within the spiral), while the spiral branches
3 pixels away are still isolated. \textbf{H}: the thin diagonal lines
behave similarly than vertical and horizontal lines. Lines too close
at either scale are merged into a unified texture with a surrounding
edge.

This example clearly demonstrates the effect on changing the spatial
scale $\lambda$. The next section also shows the effect of changing
the data scale $\kappa$, on a real image presenting both low and
large gray scale contrasts.

\subsection{\label{sub:benchmark}Typical behavior of the algorithm on a real
image}

\begin{figure}
\includegraphics[width=1\columnwidth]{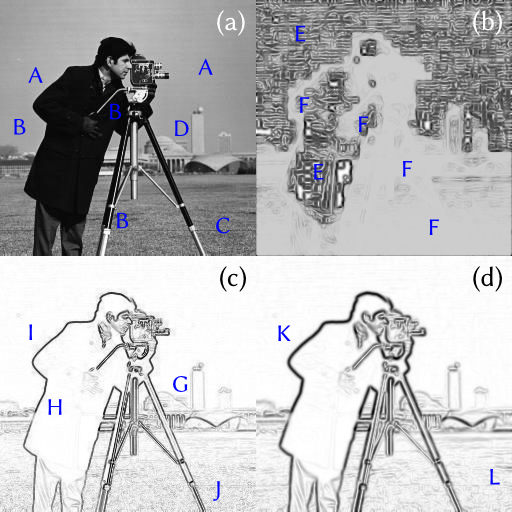}

\caption{\label{fig:cameraman_results}Scale-dependent edge detection. a. Original
$256\times256$ gray scale cameraman image. b. Analysis for $\lambda=3$,
$\kappa=1/256$. c. Analysis for $\lambda=1$, $\kappa=1$. d. Analysis
for $\lambda=3$, $\kappa=1$. Capital letters markers are commented
in the main text.}
\end{figure}

The effect of analyzing the image at different spatial and data scales
is demonstrated is Fig.~\ref{fig:cameraman_results}. The classic
``cameraman'' picture is shown in Fig.~\ref{fig:cameraman_results}a.
This picture presents jpeg quantization artifacts in the sky region
and around the coat (\textbf{A}), large contrasts for the coat, the
tripod and the camera handle (\textbf{B}), and fine grass texture
with intermediate contrasts (\textbf{C}). Buildings in the background
form another medium contrast set (\textbf{D}). At a very low $\kappa=1/256$
of one gray level quantization (Fig.~\ref{fig:cameraman_results}b),
the STD is sensitive to small changes at that level. Boundaries are
detected on jpeg artifacts and within the coat (\textbf{E}). On the
other hand, all medium and large gray level gradients $\gg\kappa$
are ignored, as can be seen on the coat edge, camera handle and tripod
(\textbf{F}) that have completely disappeared in this picture. The
grass texture at intermediate gray level contrasts is also ignored.
Conversely, setting $\kappa=1$ in Fig.~\ref{fig:cameraman_results}c
and Fig.~\ref{fig:cameraman_results}d, results in edges that are
sensitive to large gray level differences, with the reversed situation
of ignoring small gray-level discrepancies. Buildings with intermediate
contrast result in intermediate edge values (\textbf{G}). At $\lambda=1$
pixel in Fig.~\ref{fig:cameraman_results}c, edges are thin (\textbf{H})
but some jpeg quantization blocs are still faintly visible around
the coat (\textbf{I}). This spatial scale is too low for matching
the grass texture (\textbf{J}). With a larger $\lambda=3$ pixels
in Fig.~\ref{fig:cameraman_results}d the edges become smoother (\textbf{K}),
but jpeg artifacts are completely gone. Details of the texture finer
than $\lambda=3$ are merged (\textbf{L}).

The ability of an STD analysis to focus on specific scales could be
very useful, for example, as a filtering guide for artifact removals
(low $\kappa$) or for implementing new edge detectors (large $\kappa$).

\subsection{\label{sub:SST}Use on physical measuments with characteristic scales}

\begin{figure}
\includegraphics[width=0.495\columnwidth]{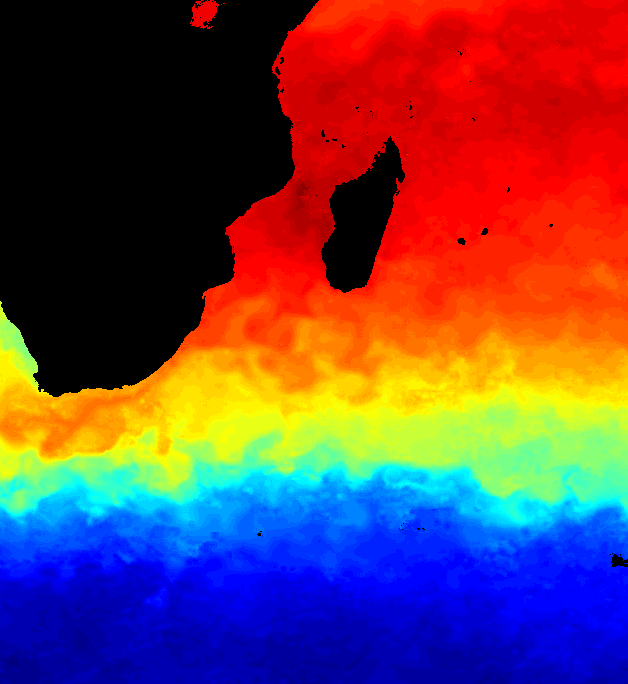}\hfill{}\includegraphics[width=0.495\columnwidth]{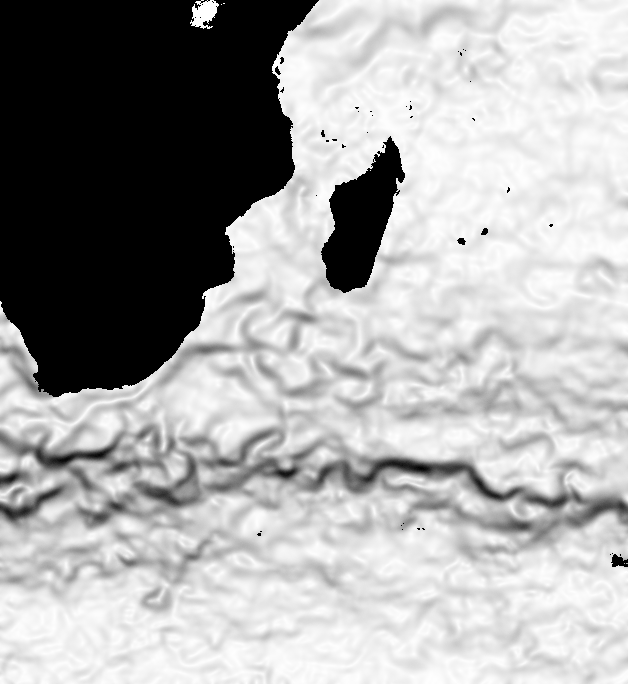}

\caption{\label{fig:Sea-surface-temperature.}Sea surface temperature. Left:
MODIS data, dated 01/01/2006, covering 15\textdegree{}E to 70\textdegree{}E
in longitude and 0\textdegree{}S to 60\textdegree{}S in latitude in
$628\times684$ pixels. The floating-point temperature is shown in
color scale from blue (-1.2\textdegree{}C) to red (31.5\textdegree{}C).
Missing values are shown in black. Right: Analysis with $\kappa=1$\textdegree{}C
and $\lambda=75$km (at the center of the picture).}
\end{figure}

The ``cameraman'' example demonstrates the ability of the algorithm
to analyze and highlight image features at prescribed spatial and
data scales. This is especially useful when dealing with physical
data, for which processes occur at known scales, independently of
the sampling resolution by which that process is measured (i.e. $\lambda$
is better expressed in real units rather than in pixels). As a demonstration,
Fig.~\ref{fig:Sea-surface-temperature.}left shows remotely sensed
sea surface temperature (s.s.t.), acquired in the infrared band with
the MODIS instrument aboard the Terra and Aqua satellites. The covered
region ranges from 15\textdegree{}E to 70\textdegree{}E in longitude
and 0\textdegree{}S to 60\textdegree{}S in latitude. We use an 8-day
composite image in order to remove the cloud cover. Missing data thus
correspond to land masses, and is dealt with as described in Section~\ref{sub:Handling-missing-data}.
The floating-point values of the temperature are used as input, from
-1.2\textdegree{}C to 31.5\textdegree{}C. The processes we are looking
for are oceanic currents, which typically induce a temperature variation
of at most a few degrees, so we take a characteristic data scale of
$\kappa=1$\textdegree{}C. Their width is variable, between 50-100km,
with transitions zones to the surrounding water of a few km, so we
take an a priori spatial scale of $\lambda=75$km. Fig.~\ref{fig:Sea-surface-temperature.}right
shows the result of the analysis of the sea surface temperature with
these parameters, which clearly highlights the dynamics of the ocean
at the prescribed scales.

\section{\label{sec:Multi-scale-analysis}Multi-scale analysis}

\subsection{Method}

\begin{figure*}
\includegraphics[width=1\textwidth]{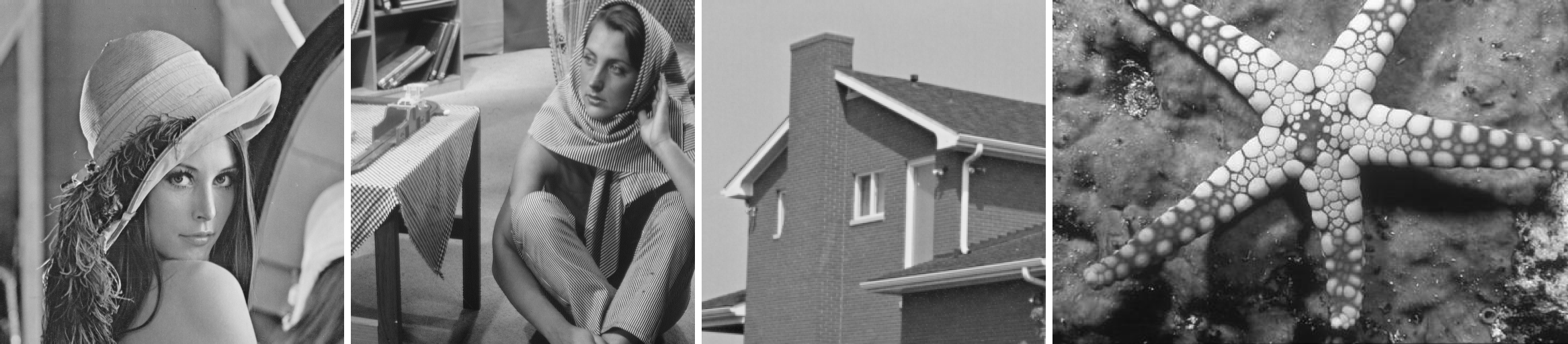}

\caption{\label{fig:Reference-images}Classic reference images. From left to
right, lena, barbara, house and sea star. We use the versions of lena,
barbara and house from \cite{portilla03}. The sea star picture is
the training example 12003 from the Berkeley segmentation data set
\cite{berkeley_segmentation}. We choosed it as it presents some background
texture with higher intensity spots, some blurry and some sharp edges.}
\end{figure*}
\begin{figure*}
\includegraphics[width=1\textwidth]{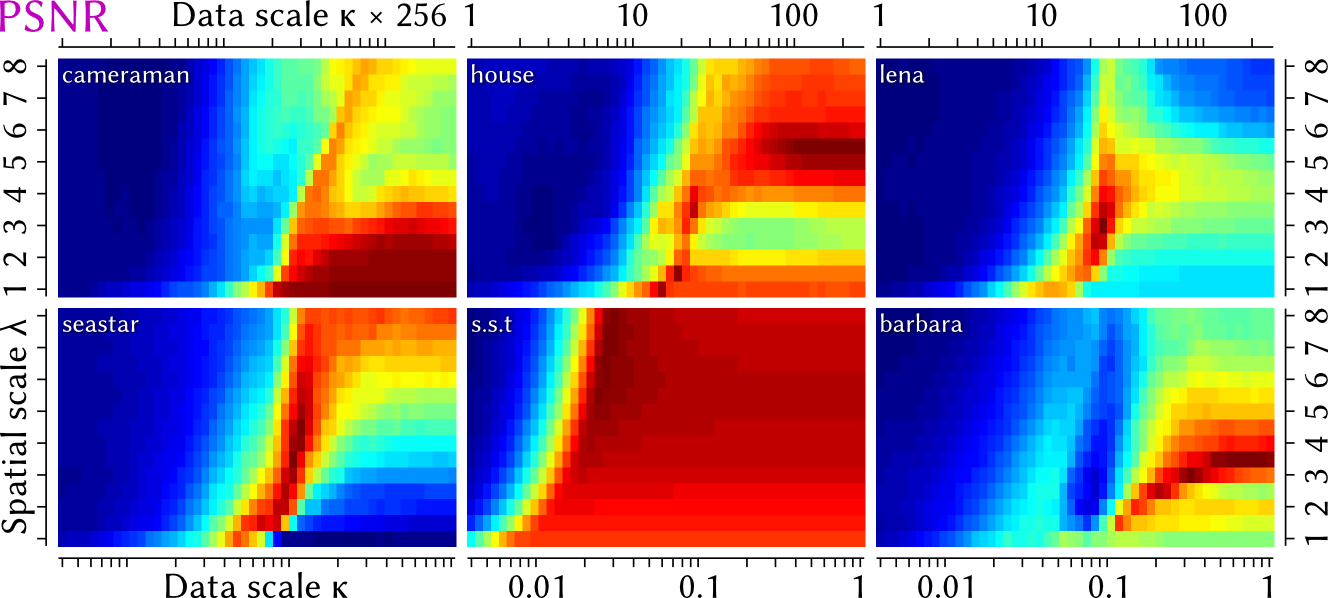}

\caption{\label{fig:psnr}Influence of the two scale parameters $\lambda$
and $\kappa$ on the reconstruction accuracy, measured with the peak
signal to noise ratio (PSNR) as defined in the main text, when retaining
20\% of the points with higher STD. Averages of 5 independent computations
with $n=100$ were used for computing the PSNR. The minimum and maximum
values for the PSNR are given in Table \ref{tab:Reconstruction-accuracy},
for each image. Using these per-image bounds, blue is the lowest accuracy,
red the highest. The spatial scale $\lambda$ is given in pixels,
the data scale $\kappa$ is given for $[0..1]$-normalized data values.
An additional scale $\kappa\times256$ is shown on the top to ease
the interpretation of gray-scaled image results. Conversion factors
for the sea surface temperature data are $\kappa\approx0.03$ for
$1$\textdegree{}C and $\lambda\approx1$ pixel for $8.46$ km at
the center of Fig.~\ref{fig:Sea-surface-temperature.}.}
\end{figure*}
The previous section presents the case where scales are chosen a priori,
according to the user's knowledge or objectives. But such information
is not always available: in many cases, the goal is precisely to find
what are the relevant scales to describe a data set. We thus need
a methodology to identify which are the ``best'' scales in an image,
with objective measurement criteria.

One hypothesis is that the ``best'' scales should identify correctly
the most relevant structures in a data set, which we identify with
the structures carrying most of the information. Assuming this is
the case, then we propose the following methodology:

\textendash{} Analyze the data for a given pair of scales $\lambda$
and $\kappa$. 

\textendash{} Select the 20\% most discriminative points, taken to
be pixels $\boldsymbol{x}$ with the largest $d(\boldsymbol{x})$
STD values.

\textendash{} Reconstruct the image from only these points, according
to the method detailed below.

\textendash{} Compare the reconstructed image with the original.

If the features are correctly determined and they indeed carry most
of the information on the image, then the reconstruction should be
very close to the original. This procedure can also be seen as a lossy
compression of an image: information in the discarded pixels is lost
and these are reconstructed from the retained pixels. Perfect features
holding all the information would mean perfect restoration of the
image after decompression. We emphasize that our goal is not to obtain
the largest PSNR from the minimum number of retained points, which
is related topic adressed by other works in the litterature \cite{nonlocal_restoration}.
Rather, we wish to use the simplest method, so that the reconstruction
accuracy measures the sole influence of $\lambda$ and $\kappa$,
and not their joint effect with reconstruction methods designed to
improve the PSNR. This excludes anisotropic diffusion \cite{anisotropic_msm_reconstruction},
weighting \cite{weighted_poisson}, exploiting dictionaries \cite{ksvd_dico},
etc.

Reconstruction is thus performed by minimizing a least square error,
equivalent to solving a Poisson Equation \cite{poisson_eq,image_reconstruction}.
More precisely, considering all pairs of neighbor valid pixels $\boldsymbol{a}$
and $\boldsymbol{b}$ (i.e. both are within image boundaries and without
missing data), we seek the reconstructed image $\boldsymbol{J}$ with
pixel values $v_{\boldsymbol{J}}\in\mathcal{V}$ such that:

\begin{equation}
\boldsymbol{J}=\left\{ v_{\boldsymbol{J}}:\,\min\sum_{\boldsymbol{a},\boldsymbol{b}}\left(v_{\boldsymbol{J}}(\boldsymbol{b})-v_{\boldsymbol{J}}(\boldsymbol{a})-g(\boldsymbol{a},\boldsymbol{b})\right)^{2}\right\} \label{eq:reconstruction}
\end{equation}

Here $g(\boldsymbol{a},\boldsymbol{b})=v_{\boldsymbol{I}}(\boldsymbol{b})-v_{\boldsymbol{I}}(\boldsymbol{a})$
is the gradient of the original image $\boldsymbol{I}$ whenever pixel
$\boldsymbol{a}$ is retained amongst the 20\% with largest STD value,
and null otherwise. This reconstruction is unique, optimal in the
least square sense, and does not introduce any extra processing step
on the image. It can be computed simply with sparse optimizers, such
as \cite{Ceres}. We then add the mean of the original image that
was lost in the process before comparing $\boldsymbol{J}$ and $\boldsymbol{I}$,
and clamp pixels within the normalized $[0,1]$ interval. 

In order to assert the quality of the reconstruction, we use the Peak
Signal to Noise Ratio (PSNR) criterion, as defined by:
\[
\textrm{PSNR}\left(\boldsymbol{I},\boldsymbol{J}\right)=10\log_{10}\left(\left|C\right|/\sum_{\boldsymbol{a}\in C}\left(v_{\boldsymbol{I}}(\boldsymbol{a})-v_{\boldsymbol{J}}(\boldsymbol{a})\right)^{2}\right)
\]
 where $C$ is the set of common valid pixels in $\boldsymbol{I}$
and $\boldsymbol{J}$.

\subsection{Results}

In addition to the cameraman and sea surface temperature images, we
also analyze the ``lena'', ``barbara'', ``sea star'' and ``house''
images, shown in Fig.~\ref{fig:Reference-images}. Their multiscale
analysis maps are shown in Fig.~\ref{fig:psnr} and the PSNR maxima
are given in Table \ref{tab:Reconstruction-accuracy}. Scales leading
to better global reconstructions are clearly visible, forming distinct
patterns for every picture.

\begin{table}
\caption{\label{tab:Reconstruction-accuracy}Reconstruction accuracy when retaining
20\% of the points and best scales for the test images.}

\centering{}%
\begin{tabular}{|c|c|c|c|c|}
\hline 
Image & Best $\lambda$ & Best $\kappa$ & min PSNR & max PSNR\tabularnewline
\hline 
cameraman & 1 & 0.41 & 12.0 & 28.3\tabularnewline
\hline 
house & 5.5 & 0.64 & 13.5 & 26.2\tabularnewline
\hline 
lena & 3 & 0.094 & 13.6 & 21.0\tabularnewline
\hline 
seastar & 3.5 & 0.11 & 13.7 & 18.2\tabularnewline
\hline 
s.s.t. & 7 & 0.027 & 10.8 & 23.9\tabularnewline
\hline 
barbara & 3.5 & 0.64 & 13.0 & 17.7\tabularnewline
\hline 
\end{tabular}
\end{table}

Our method provides a new multi-scale analysis of an image, in order
to highlight the characteristic spatial $\lambda$ and data $\kappa$
scale of its components. For example, we can see that the house picture
has a maxima zone at $\lambda=5$ to $6$ and high $\kappa$: this
corresponds to the spatial extent and the contrast difference of the
white borders of the roof, window frames and water drain. Other maxima
are located at $\lambda$ of $1$ and $1.5$ pixels, with medium $\kappa$
gray scale contrast: these are the brick textures. With this method,
automatically identified parameters for the sea surface temperature
data are $\lambda\approx59$km and $\kappa\approx0.9$\textdegree{}C.
These are consistent with our a priori estimate in Sec.~\ref{sub:SST},
further validating our approach.

\subsection{\label{sub:Selective-texture-erasing}Selective texture erasing}

\begin{figure*}
\includegraphics[width=1\textwidth]{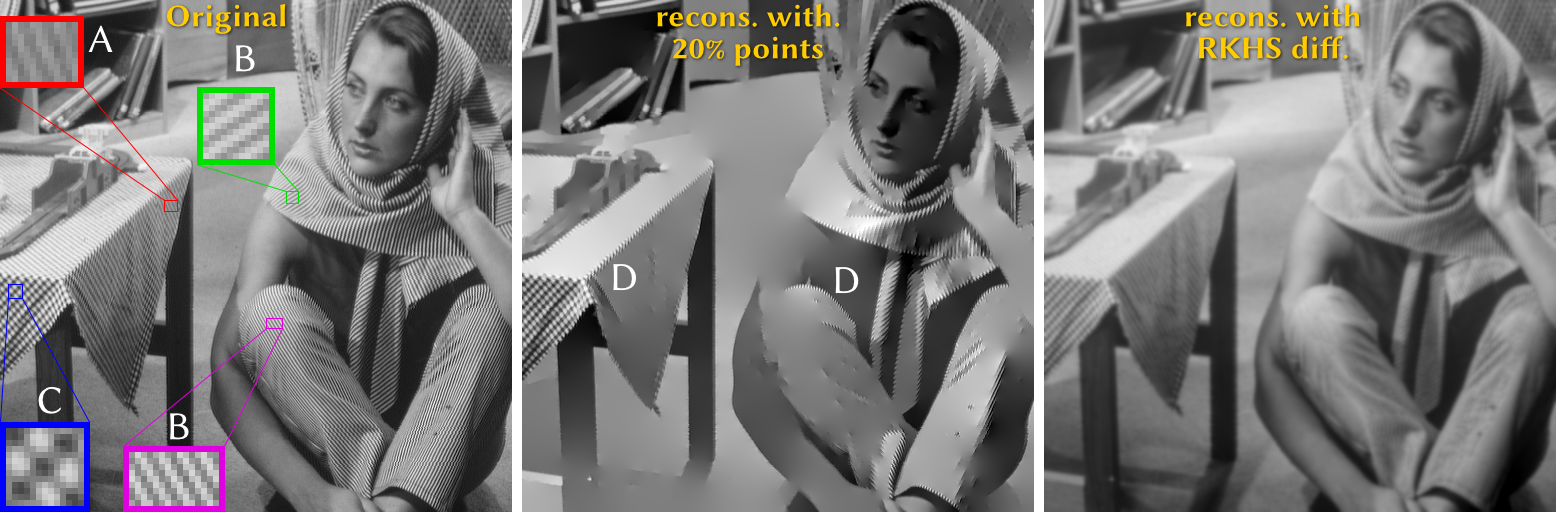}

\caption{\label{fig:recbarbara}Erasing texture elements smaller than $\lambda=3.5$
pixels. Left: Annotated original Barbara image. Middle: Reconstruction
from 20\% selected points at $\lambda=3.5$ and $\kappa=0.64$, corresponding
to the maximum of Table \ref{tab:Reconstruction-accuracy}. Right:
Reconstruction with 100\% points but using the RKHS ``texture gradient''
instead of the image gradient.}
\end{figure*}

Similarly, the stripped patterns of the Barbara image are correctly
matched on both scales. Fig.~\ref{fig:recbarbara} shows in region
\textbf{A} a texture of slanted stripes with a characteristic size
$\lambda<2.5$ and a low gray level constrast $\kappa$. These textures
are correctly detected in the PSNR map of Fig.~\ref{fig:psnr} as
the leftmost maxima with lower $\lambda$ values. But the image also
comprises other patterns of increasing contrast and size, as can be
seen in regions \textbf{B}. Most of them have a characteristic size
$\lambda=3.5$, at which the PSNR reaches a maxima for a range of
large gray scales $\kappa$.

Fig.~\ref{fig:recbarbara} (middle) shows that textures disappear
in the reconstructed image from the 20\% largest STD pixels. Some
patterns, like in region \textbf{C}, present textures with a larger
spatial extent. As was the case for the artificial image in section
\ref{fig:refpics}, these patterns are not recognized with a smaller
$\lambda=3.5$, and they appear intact in the reconstruction. Our
method can thus also be used for texture erasing \cite{poisson_editing,image_smoothing},
preserving edges between different regions and around textured elements
(\textbf{D}) but selectively removing all textures below a prescribed
spatial scale $\lambda$.

An alternative method is to seek the image with gradients (Eq.~\ref{eq:reconstruction})
that are as close as possible to the texture difference values (Eq.~\ref{eq:MMD}).
Since there cannot be a linear correspondance between pixel gradients
in image space and ``texture gradients'' in RKHS, involving a non-linear
reproducing kernel, it is expected that distortions are produced in
the reconstructed image, assuming a meaningful image can be produced
in the first place. Technical details for how to combine Eq.~\ref{eq:MMD}
into Eq.~\ref{eq:reconstruction} are given in Appendix C, and involve
orienting the ``texture gradient'' with the average intensity on
each side of a target pixel. Despite being necessarily imperfect,
Fig.~\ref{fig:recbarbara} (right) shows that a reconstruction can
still be obtained this way, with a PSNR of 22.95 and erased textures
below $\lambda$, but with some amount of blurring. The interesting
point is that the method works as intended, proving that our ``texture
gradient'' operation in RKHS, when properly oriented, really behaves
like a gradient.

\section{\label{sec:Color-texture-differences}Color texture differences}

\begin{figure*}
\includegraphics[width=1\textwidth]{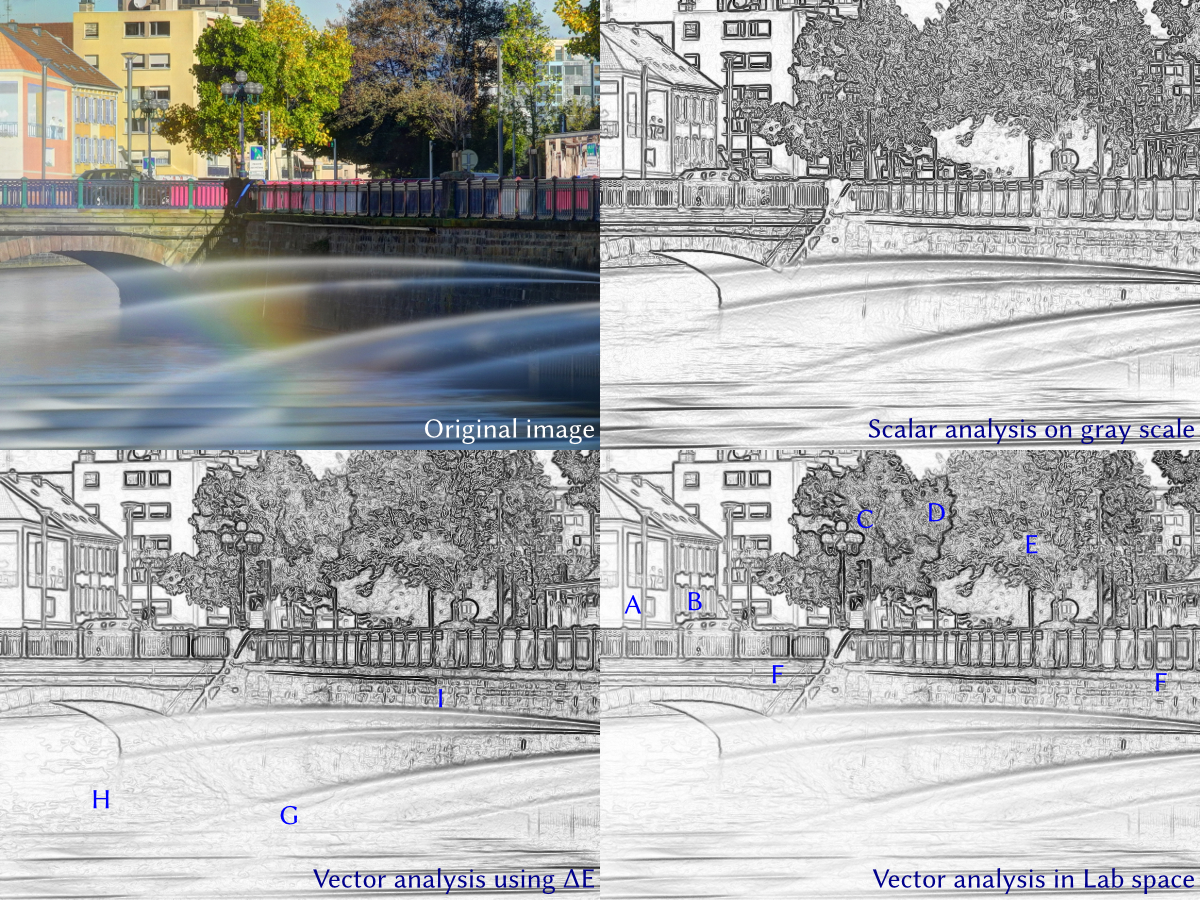}

\caption{\label{fig:belfort}Texture difference analysis of a color image.
Top-left: original image (full credit given in Appendix), rescaled
to 600x450 pixels. Top-right: Scalar analysis using a gray version
of the image. Bottom-right: Vector analysis in Lab space with the
standard D65 white point. Bottom-left: Vector analysis using the CIE
DE2000 perceptual color difference improvement $\Delta E$. All these
analyses use the spatial scale $\lambda=1.5$ pixels and the data
scale $\kappa=0.15$.}
\end{figure*}

The path generation procedure described in Section \ref{sub:Comparing-two-sequences}
makes no assumption on the nature of pixel values $v\in\mathcal{V}$,
just that sequences $t\in\mathcal{V}^{m}$ can be compared with a
suitable reproducing kernel. The case for scalar pixels is presented
above, but the method works equally well on vector data. In this case,
$\mathcal{V}$ could be a color space, a set of spectral bands in
remote sensing applications, polarized synthetic aperture radar values,
or any other vector data. We deal here with the most common case of
a triplet of RGB values. We first consider a kernel $k:\,\mathcal{V}\times\mathcal{V}\rightarrow\mathbb{R}$
that compares two RGB triplets and produce a similarity value, then
take the product space kernel in $\mathcal{V}^{m}$ for comparing
sequences.

Specifically, assuming $\mathcal{V}$ is the set of all $v=\left(R,G,B\right)$
triplets, we first consider the non-linear conversion operator $\ell:\,\mathcal{V}\rightarrow\mathcal{L}$
to the CIE Lab space $\mathcal{L}$ using the standard D65 white point.
$\ell(v)$ is thus a vector with entries $\left(L,a,b\right)$. Then,
we implement the following color difference operators:

\textendash{} $\delta_{1}\left(v,w\right)=\left\Vert \ell\left(v\right)-\ell\left(w\right)\right\Vert _{\mathcal{L}}/100$,
based on the original intent of the Lab space to be perceptually uniform,
so the norm in $\mathcal{L}$ should match a perceived color difference.
This first formula is fast to compute.

\textendash{} $\delta_{2}\left(v,w\right)=\Delta E\left(\ell\left(v\right),\ell\left(w\right)\right)/100$,
based on the revised CIE DE 2000 standard \cite{CIEDE2000} for producing
a better perceptually uniform difference $\Delta E$ between two Lab
triplets. This second formula is slower to compute, but presumably
more accurate.

The kernel between two sequences $s,t\in\mathcal{V}^{m}$ is then
easily adapted from Eq.~\ref{eq:kernel_with_NaN}, with the same
notations:

\begin{equation}
k_{d}\left(s,t\right)=1/\left(1+\frac{1}{\left|C\right|}\sum_{i\in C}\left(\delta_{d}\left(s_{i},t_{i}\right)/\kappa\right)^{2}\right)\mbox{ with }d=1,2\label{eq:color_kernel}
\end{equation}

The data scale $\kappa$ is now applied to the chosen $\delta_{1}$
or $\delta_{2}$ operator, in order to highlight small or large color
differences. Note that we normalize $\delta_{1}$ and $\delta_{2}$
to be within $\left[0\ldots1\right]$ instead of the $\left[0\ldots100\right]$
Lab space range, so that $\kappa$ values for the color case are comparable
to the scalar case presented in the previous sections: assuming the
gray scale perfectly matches a perceptual color intensity, then $\kappa$
would have the same meaning in the color and the gray scale analysis%
\footnote{In practice, this is not exactly the case, and gray scale conversion
is an active topic of research \cite{grayconv}. Here we use the Y
component of XYZ space, which is also used by The Gimp free software
default's gray conversion tool.%
}. Using the same normalized $\kappa$, the scalar and vector analysis
in Fig.~\ref{fig:belfort} are thus expected to have a similar, but
not exactly identical, general contrast range. On the other hand,
the CIEDE2000 $\Delta E$ formula was designed to correct slight discrepancies
in the original Lab space perceptual uniformity, hence the contrast
levels of the two vector analysis should be very similar, with only
minor changes due to the $\Delta E$ correction. This is what we observed
in Fig.~\ref{fig:belfort}.

Fig.~\ref{fig:belfort} shows how the vector analysis, using either
the Lab space norm or $\Delta E$ for color matching, enhances the
detected features compared to the scalar analysis on gray scale. In
region \textbf{A}, the color difference in the top-left of the trompe-l'œil
painting is well recognized in the vector analysis, but the contrast
is not large enough in gray scale. The real wall and windows on the
same house (\textbf{B}) are also much better analyzed in color, especially
with the orange/beige difference. Another prominent difference is
the lamp post (\textbf{C}) in front of the tree, which is lost in
foliage details in the gray scale version but clearly contoured in
the color analysis. Unlike points \textbf{A} and \textbf{B}, this
effect is not only due to color contrast, but it also involves the
texture difference between the lamp post and the tree foliage. Texture
differencing between the two trees is clearly visible in \textbf{D},
where the green tree is well contoured in the color analysis, while
both trees have similar textures in gray scale. The small patchs of
blue sky (\textbf{E}) better stand out against more uniform foliage
textures in the color versions, compared to the gray case where they
are buried in noise. Color panels are also better recognized in \textbf{F}.
The rainbow is not detected (\textbf{G}) by any of the analyses. But
its slowly varying colors over a large spatial extent do not match
the scales $\lambda=1.5$ pixels and $\kappa=0.15$ that are used
here.

The effect of using the CIEDE2000 correction is visible by close inspection
of the bottom images in Fig.~\ref{fig:belfort}. Compared to the
original Lab space, the non-linear $\Delta E$ correction gives slightly
more contrast to the lower color differences, which was one of its
purposes in the first place \cite{CIEDE2000}. These effects are visible
throughout the image, but maybe more clearly seen on the water (\textbf{H})
and stone wall (\textbf{I}) texture details. However, these effects
are so small that, for most applications, computing $\Delta E$ may
not be worth the added cost compared to working in the original Lab
space.

In any case, our method provides a new way to include color information
in image and texture analysis, highlighed especially by points \textbf{A,
B} and \textbf{C} above. More generally, the method is applicable
whenever a reproducible kernel can be defined to compare pixel elements,
whether these are scalar, vector, strings, graphs, etc.

\section{Conclusion}

We introduce a new low-level image analysis method, based on statistical
properties within the neighborhood of each pixel. We have demonstrated
its use on synthetic and real images. In particular, our algorithm
is able to retreive characteristic scales in remotely sensed data.
The method we present is not limited to scalar values and it is readily
applicable to any kind of data for which a reproducing kernel is available.
This includes vector data, and in particular color triplets. Thanks
to being specified at prescribed spatial and data scales, our method
implements a new kind of filter, very different from wavelet-based
analysis \cite{Starck02} or decomposition on dictionaries \cite{Mairal08}.
For example, it can detect small discrepancies like jpeg quantization
artifacts while being insensitive to large luminosity gradients, which
could be useful for guided filtering of these artifacts. In the spatial
domain, our method can detect repetitive texture patterns, as shown
in Figs. \ref{fig:refpics} and \ref{fig:recbarbara}, and produces
a unique contour around these elements. This low-level algorithm represents
textures in a statistical framework that is quite different from classic
approaches like textons \cite{Malik01}, and it also represents multi-scale
information in a new framework. Our new method thus complements these
techniques, and together with them has the potential to provide really
new feature descriptors for images, with properties that need to be
explored in future works.

\section*{Appendix A: Producing texture discrepancy for missing data}

For directions tangent to an invalid data zone, half the neighborhood
is valid and half is invalid, in both $S^{+}$ and $S^{-}$(at worst,
since $\boldsymbol{N}^{\pm}(\boldsymbol{o})$ is centered on the center
of the valid pixel $\boldsymbol{o}$, not on the edge). For such a
$S^{\pm}$ pair, and replacing the neighborhood by another worst-case
approximation of a binary valid/invalid choice (i.e. neglecting $m$
and the spatial extent $\lambda$ with many valid pixels in the neighborhood),
the probability of never selecting a valid pair of pixels decreases
exponentially as $O(2^{-n^{2}})$ according to Eq. \ref{eq:MMD},
tending to null probability in the large $n$ limit. For a typical
computation with $n=100$, this a gives ridiculously small $p<10^{-3000}$
probability of an invalid result. Thus, for all practical purposes,
valid pixels near an invalid boundary ``always'' produce a valid
computation result for typical and small values of $n$.

In fact, even missing pixels within large missing data zones benefit
from nearby valid pixels. Although we explicitly ignore these STD
values due to their low accuracy, our method still allows getting
relevant values for isolated missing pixels. Consider a neighborhood
centered on that pixel. In that case, paths around that point may
occasionally fall onto the missing pixel, but patterns are still matched
outside that pixel. Since the modified kernel is normalized only on
valid pixels, the missing data simply results in a reduction of the
effective $m=\left|C\right|$, but valid patterns surrounding the
missing pixel are still matched correctly in every direction. The
price to pay is a loss of reliability, especially as the nearest distance
to a valid pixel grows. We do not exploit this feature in the data
presented in the main text, where no STD value is produced for missing
pixels, but this feature could be useful in other contexts, for example
to simply ignore isolated missing pixels.

\section*{Appendix B: Combining scales}

In Section \ref{sub:Choosing_kappa}, $\kappa$ sets the scale at
which scalar pixel values are compared and should reflect a priori
information on the image. In order to compare probability distributions,
\cite{RKHS_proba_measures} proposes to either take a Bayesian approach
for letting $\kappa$ vary with a priori knowledge, or to take $d\left(S^{-},S^{+}\right)=\sup_{\kappa}\, d_{\kappa}\left(S^{-},S^{+}\right)$
over a range of $\kappa$ for a more systematic approach. We found
that the latter does not perform well in practice. For example, a
picture captured by a digital camera sensor in low light conditions
presents some digital noise on the pixel values. A similar effect
occurs for quantization noise and jpeg artifacts (see Section \ref{sub:benchmark}).
Comparing distributions in $\mathcal{V}^{m}$ at the scale of this
noise is meaningless, as it would reflect differences due to the sensor
(resp. quantization algorithm) itself, but not the differences in
the image textures. Given the sensor fluctuations, the distance $d_{\kappa}\left(S^{-},S^{+}\right)$
at low $\kappa$ values is larger that at the natural scale of the
image signal, hence the $\sup_{\kappa}$ approach does not work in
this case. Physically, and for remotely sensed data in particular,
the $\sup_{\kappa}$ approach amounts to mixing together processes
at different characteristic scales, which may have nothing to do with
each other.

\section*{Appendix C: Reconstruction from ``texture gradients''}

For a pair of valid neighbor pixels $(\boldsymbol{a},\boldsymbol{b})$,
consider the neighborhood $\boldsymbol{N}^{+}(\boldsymbol{a})$ that
includes $\boldsymbol{b}$ and $\boldsymbol{N}^{-}(\boldsymbol{b})$
that includes $\boldsymbol{a}$. We add constraints in Eq.~\ref{eq:reconstruction}
using Eq.~\ref{eq:MMD} for the gradients: $\min\sum_{\boldsymbol{a},\boldsymbol{b}}\left(v_{\boldsymbol{J}}(\boldsymbol{b})-v_{\boldsymbol{J}}(\boldsymbol{a})\pm d\left(S^{-}(\boldsymbol{a}),S^{+}(\boldsymbol{a})\right)\right)^{2}+\left(v_{\boldsymbol{J}}(\boldsymbol{b})-v_{\boldsymbol{J}}(\boldsymbol{a})\pm d\left(S^{-}(\boldsymbol{b}),S^{+}(\boldsymbol{b})\right)\right)^{2}$.
The signs are determined using the average value of the original image
pixels in each neighborhood: $\mbox{\textrm{sign}\,}d\left(S^{-}(\boldsymbol{a}),S^{+}(\boldsymbol{a})\right)=\textrm{sign}\left(\sum_{\boldsymbol{x}\in\boldsymbol{N}^{+}(\boldsymbol{a})}p(\boldsymbol{x})v_{\boldsymbol{I}}(\boldsymbol{x})-\sum_{\boldsymbol{x}\in\boldsymbol{N}^{-}(\boldsymbol{a})}p(\boldsymbol{x})v_{\boldsymbol{I}}(\boldsymbol{x})\right)$,
and similarly for $\boldsymbol{b}$. With this method, the result
for $\lambda=3.5$ and $\kappa=0.64$ is shown in Fig.~\ref{fig:recbarbara}
right.

\section*{Appendix D: Data and code availability}

The code used in this paper is free software, available for download
at \url{https://gforge.inria.fr/scm/?group_id=5803}. Images for Lena,
Barbara, house, were fetched from J.~Portilla's web page and match
these used in \cite{portilla03}. We used the cameraman picture from
McGuire Graphics Data at \url{http://graphics.cs.williams.edu/data/images.xml}.
The sea star is the training sample 12003 from the Berkeley segmentation
S300 data set \cite{berkeley_segmentation} and was downloaded from
\url{http://www.eecs.berkeley.edu/Research/Projects/CS/vision/bsds/BSDS300/html/dataset/images/gray/12003.html}.
The sea surface temperature data was provided by the Legos laboratory
\url{http://www.legos.obs-mip.fr/}. The color image in Fig.~\ref{fig:belfort}
comes from Wikimedia Commons, at \url{http://commons.wikimedia.org/wiki/File:2013-10-19_10-53-16-savoureuse-belfort.jpg}.
This long exposure photograph was taken by Thomas Bresson with a polarizing
filter, and released under Creative Commons - Attribution - 3.0. The
original image was cropped and resized to $600\times450$ pixels.

\section*{Acknowledgements}

We thank Hicham Badri for useful discussions and examples of image
reconstruction with the Poisson equation.

\end{document}